# AI-Powered Urban Green Infrastructure Assessment Through Aerial Imagery of an Industrial Township

Anisha Dutta*

*Data & Analytics, Tata Steel Limited
anisha.dutta@tatasteel.com

## Abstract

Accurate assessment of urban canopy coverage is crucial for informed urban planning, effective environmental monitoring, and mitigating the impacts of climate change. Traditional practices often face limitations due to inadequate technical requirements, difficulties in scaling and data processing, and the lack of specialized expertise. This study presents an efficient approach for estimating green canopy coverage using artificial intelligence, specifically computer vision techniques, applied to aerial imageries. Our proposed methodology utilizes object-based image analysis, based on deep learning algorithms to accurately identify and segment green canopies from high-resolution drone images. This approach allows the user for detailed analysis of urban vegetation, capturing variations in canopy density and understanding spatial distribution. To overcome the computational challenges associated with processing large datasets, it was implemented over a cloud platform utilizing high-performance processors. This infrastructure efficiently manages space complexity and ensures affordable latency, enabling the rapid analysis of vast amounts of drone imageries. Our results demonstrate the effectiveness of this approach in accurately estimating canopy coverage at the city scale, providing valuable insights for urban forestry management of an industrial township. The resultant data generated by this method can be used to optimize tree plantation and assess the carbon sequestration potential of urban forests. By integrating these insights into sustainable urban planning, we can foster more resilient urban environments, contributing to a greener and healthier future.

**Contents.**

## I.    Introduction

Green canopy coverage, a key metric for urban planning and environmental assessment, refers to the proportion of land covered by the vertical projection of tree crowns. Traditionally, this measurement was a laborious process, requiring dedicated personnel to manually calculate canopy coverage from aerial photographs or maps using analogue tools, and even manual inspection. This approach was not only time-consuming but also prone to human error, hindering the efficiency and accuracy of urban green space analysis. However, recent advancements in artificial intelligence, specifically computer vision techniques, offer promising new avenues for efficient and accurate canopy assessment, sustainably paving the way for data-driven urban planning and management.

Jamshedpur is a large city, set between the Subarnarekha and Kharkai rivers in the state of Jharkhand. Right from the onset, Jamshedpur, arguably India's first planned city, was laid out according to the founder, JN Tata's idea of a town with wide streets planted with shady trees, plenty of space for lawns and gardens.[2] But, being an industrial hub, like other business cities, Jamshedpur faces constant challenges of maintaining a balance between man, machine, and nature. Therefore, for a city like Jamshedpur, the green canopy coverage calculation is indeed crucial for urban planning, environment monitoring, and climate change mitigation. Green canopies provide shades, absorbs carbon dioxide, and reduces urban heat island effects. The resulting percentage of green canopy coverage helps the urban planners, policymakers, and government officials to make informed decisions about urban forestry management, garden and park development, and sustainability initiatives.

## II. Jamshedpur City Image Data

Green canopy coverage calculation primarily needs the image data of the urban city or the land of interest. Traditional methods often rely on labor-intensive manual measurements or analysis of remote sensing data (satellite imagery), both of which are time consuming, resource intensive. In contrast, aerial imagery, which is much cost and labor efficient to collect through drone, provides valuable insights into the extent, composition, and spatial distribution of green cover.

### A. Data Collection

A dozen aerial imagery data were provided, each capturing data from a different slice of the Jamshedpur city. All the 12 image data were in Tiff format, each having 7 bands. Each of the bands represent a unique aerial perspective of the city, varying in altitude and resolution. The images are hefty, usually varying in size from 7 GBs to 48GBs each. The 12 tiff images, namely A1 to A4, B1 to B4 and C1 to C4, are attained using OSGeo GeoTIFF Library [3] to read the data inside and get transformed into executable image format. In the following image, one can visualize the 12 tiff files forming the aerial view of the Jamshedpur City.

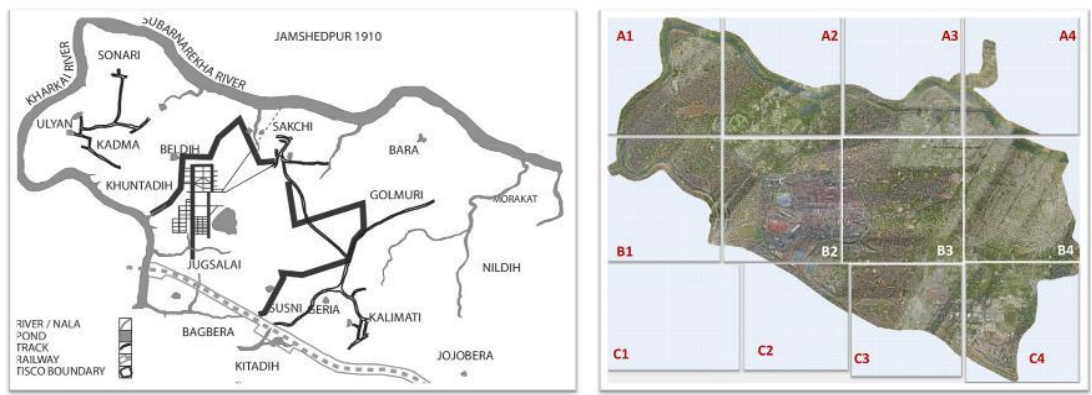

*Figure 1. Broader Spectrum View of Jamshedpur City*

### B. Data Processing

The utilization of aerial imagery in the realm of deep learning presents a unique set of challenges, particularly when it comes to data format compatibility and processing. While aerial images are often captured in the TIFF format, this format is not directly compatible with the requirements of deep learning models. Therefore, a crucial preprocessing step is required to transform these images into executable chunks that can be effectively utilized by the deep learning algorithms.

The GeoTIFF format offers a solution to this challenge. GeoTIFF, an extension of the standard TIFF format, incorporates georeferencing information, or metadata, embedded as tags within the TIFF file. This metadata provides crucial spatial referencing data, enabling the accurate location and alignment of image features within a geographic context. Leveraging the GeoTIFF library, we can access and extract the relevant data from the TIFF format files. This data is then processed further using the Zarr library, a Python package designed for managing compressed, chunked, N-dimensional arrays, ideally suited for parallel computing environments.

Zarr effectively breaks down the image data into smaller, manageable chunks, facilitating parallel processing and enhancing computational efficiency. This library excels in handling large datasets, making it a valuable tool for working with the massive amount of data generated from aerial imagery. The Zarr library extracts band-wise information from the GeoTIFF files, allowing us to isolate and analyze specific spectral bands, which contain unique information about the captured scene. For optimal performance, we select the most relevant band for our analysis, in this case, band 2. The choice of band 2 is primarily driven by the fact that the data within this band has been captured from a height that aligns well with the resolution of our training dataset. This ensures consistency and accuracy in our analysis, as the data from this band provides the most relevant information for our specific task.

The extracted data from band 2 is then converted into a NumPy array, a highly versatile data structure commonly used in deep learning models. This conversion facilitates efficient processing and analysis within the deep learning framework. The models we use in this project are image-based models, meaning they expect the input image to be of a specific shape, typically (640, 640, 3). However, aerial images often have dimensions far exceeding these requirements, often reaching tens of thousands of pixels in both height and width. To address this mismatch, we employ a chunking strategy, dividing the entire image array into smaller, manageable chunks of 640 x 640 pixels. This process involves breaking the image array into smaller, manageable units that align with the input requirements of our deep learning models. For chunks that fall at the image margins and may have dimensions smaller than 640 x 640, we pad these chunks with zero values to ensure a consistent input size. This padding technique ensures uniformity across all input chunks, maintaining consistency and facilitating efficient processing within the deep learning framework.

This chunking process results in a significant number of smaller chunks, typically around 25 thousand for a typical aerial image. These individual chunks are then stored in a Google Cloud Storage (GCS) bucket, a cloud-based storage solution that offers scalable and reliable storage for vast datasets. The GCS bucket serves as a central repository for our image chunks, providing a readily accessible source of data for the deep learning model during training and execution.

This comprehensive preprocessing pipeline, involving conversion to GeoTIFF format, extraction of relevant bands using Zarr, conversion to NumPy arrays, and chunking for compatibility with deep learning models, ensures efficient data management and facilitates the seamless integration of aerial imagery into our deep learning workflow. This approach provides a foundation for leveraging the rich information contained within aerial imagery to develop advanced solutions for urban green space analysis and management.

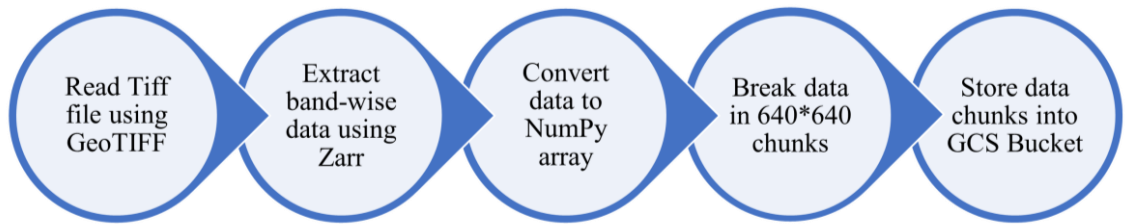

*Figure 2. Data Processing Flow*

As discussed, the images (typically the marginal chunks) with lesser height and width than 640, gets padded with zero pixels to get the shape (640, 640, 3). In the following image, one can see how a distorted chunk can be transferred to required shape without hampering the actual data of interest.

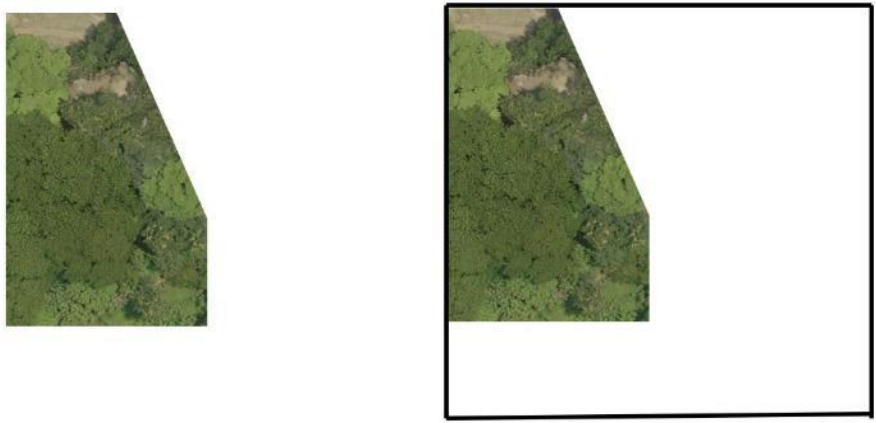

*Figure 3. [L] (Marginal) Image Chunk of random size*
*[R] Padded Image of shape (640, 640, 3)*

Since the paddings are zero values pixels, we introduce another couple of variables, namely coveredPixels and totalPixels, which keep the count for all non-zero pixels, which are basically the non-padded data. The idea is if one divides the canopy pixels by all the pixels, the resultant canopy percentage is lesser than the original one, hence we omit the padded values and divide the canopy pixels by non-padded pixels only. Also, while taking the image chunks into model execution, we ignore all those chunks where there are no non-zero values at all. This helps the whole process to be time and labor efficient. The following image explains the process pictorially. Also, we have a table which exactly displays how much 'covered' Pixels are there for each tiff image data, which can give an estimation that what percentage of the tiff data are null valued.

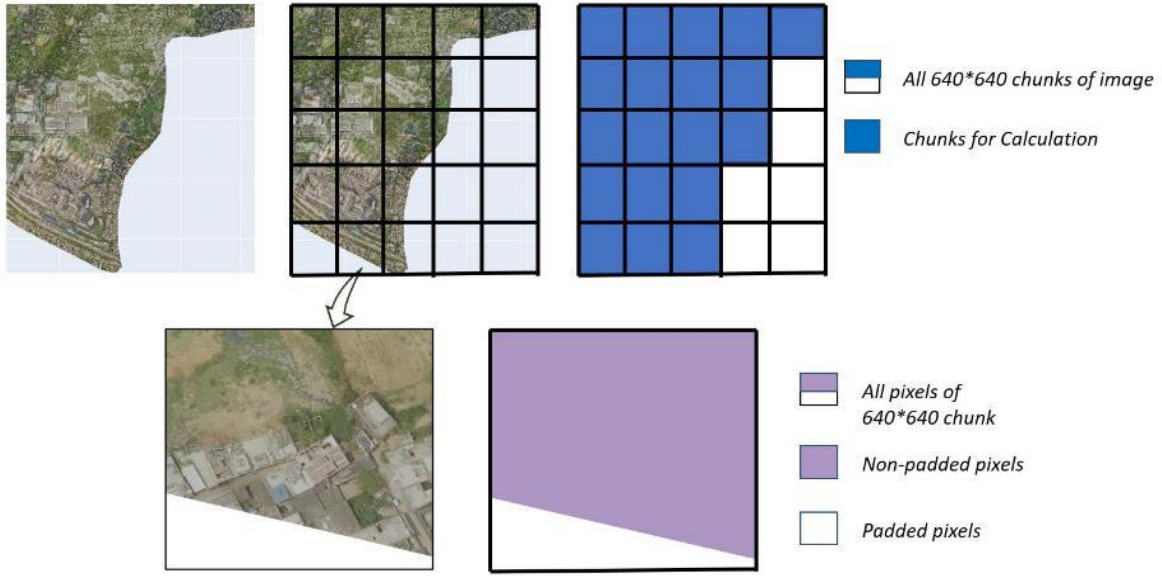

*Figure 4. [T] Gridding the converted tiff file data into 640 * 640 chunks [B] Segregation of coveredPixels from totalPixels based on padding of data*

| | fileName | totalPixels | coveredPixels | coverPercentage |
|---|---|---|---|---|
| **0** | tiles_A1 | 828993608 | 805825983 | 97.21 |
| **1** | tiles_A2 | 744243200 | 728376716 | 97.87 |
| **2** | tiles_A3 | 534528000 | 520829738 | 97.44 |
| **3** | tiles_A4 | 261734400 | 238869853 | 91.26 |
| **4** | tiles_B1 | 531251200 | 506459873 | 95.33 |
| **5** | tiles_B2 | 1296793600 | 1291256063 | 99.57 |
| **6** | tiles_B3 | 1329971200 | 1329971200 | 100.00 |
| **7** | tiles_B4 | 1172275200 | 1154281912 | 98.47 |
| **8** | tiles_C1 | 409600 | 0 | 0.00 |
| **9** | tiles_C2 | 221991680 | 212491798 | 95.72 |
| **10** | tiles_C3 | 793804800 | 776495613 | 97.82 |
| **11** | tiles_C4 | 735641600 | 710077553 | 96.52 |

*Figure 5. Percentage Calculation for Nonzero values from whole tiff data*

## III.    Canopy Coverage Calculation Approaches

Artificial Intelligence (AI) has ushered in a revolution in the world of technology over the past few decades. Its impact has permeated nearly every field, striving to make our world more reliable, cost-effective, and labor-efficient. Recent advancements in AI, particularly in the realm of deep learning algorithms, specifically Computer Vision (CV), have been instrumental in revolutionizing image processing and recognition tasks. CV models offer the potential for automated and accurate estimation of Green Canopy Coverage from street-level images, a critical component in urban planning, environmental monitoring, and climate change mitigation efforts.

Motivated by a groundbreaking research paper [1], we embarked on a study exploring the potential of Convolutional Neural Networks (CNNs) to revolutionize urban green infrastructure assessment. Our investigation led to the development of two approaches, utilizing the power of CNNs to estimate green canopy coverage from street-level imagery. These models were trained on a comprehensive dataset of street-level images meticulously annotated with ground truth canopy coverage, ensuring a robust foundation for the accurate prediction.

Prior to our main approaches, an initial attempt at calculating green percentage relied on basic computer vision tools, employing color thresholding techniques to isolate green pixels within the image. While this approach proved computationally efficient, it lacked the nuance required for accurate green canopy assessment. This method tended to misclassify non-canopy green objects, such as grassy areas or green building facades, as part of the canopy. This resulted in an inflated estimation of green coverage, highlighting the need for a more sophisticated approach that could differentiate between true canopy areas and other green elements within the urban landscape.

The first approach, which depicts image segmentation, leveraged the strengths of detailed boundary of the green canopy coverage area. This model focused on using image segmentation with a CNN-based model to accurately identify and separate green canopies from top-view images. Our second approach on the other hand, the object detection approach, employed a different strategy, focusing on identifying and locating trees in an image and highlight them with bounding boxes.

The subsequent sections of this paper delve into a comprehensive analysis of these two approaches. We meticulously examine the methodologies employed, the challenges encountered, and the comparative performance of each model. A detailed discussion of the results obtained, highlighting the strengths and limitations of each approach, provides valuable insights into their applicability and potential for real-world deployment. We also delve into the challenges we faced during the project, emphasizing the innovative solutions developed to overcome these hurdles. Ultimately, this research presents a compelling case for the transformative potential of AI-powered solutions in enhancing our understanding and sustainable management of urban green infrastructure.

## A. Segmentation Approach

Image segmentation is a computer vision technique that partitions an image into discrete groups of pixels, also called as image segments. The goal of segmentation is to simplify and/or change the representation of an image into something that is more meaningful and easier to analyze. By parsing an image's complex visual data into specifically shaped segments, image segmentation enables faster, more advanced image processing. Image segmentation techniques range from simple, intuitive heuristic analysis to the cutting edge implementation of deep learning. Conventional image segmentation algorithms process high-level visual features of each pixel, like color or brightness, to identify object boundaries and background regions. In this project, we have designed a CNN based segmentation model to aim for higher accuracy. The target was to segment the green canopies from the image itself, as shown in the following image, where is it shown that how ideally one can segment the image into green canopy segments.

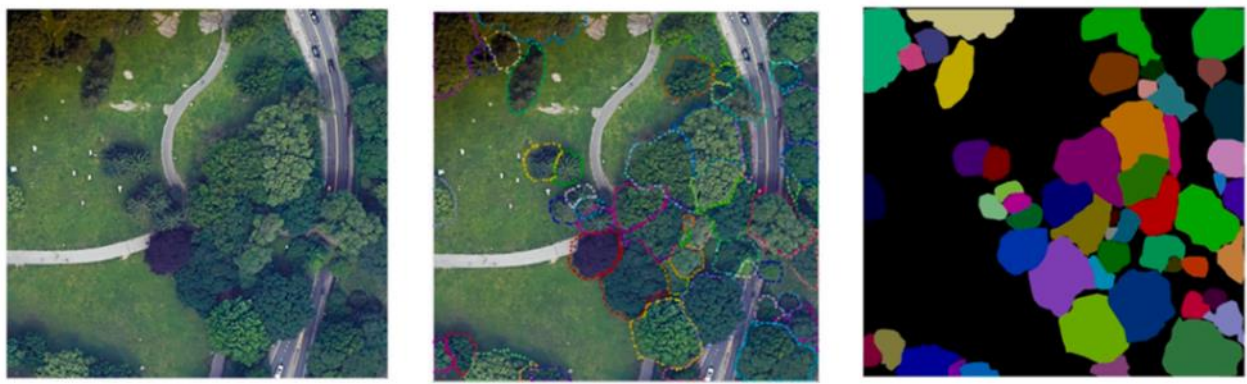

*Figure 6. Ideal Segmentation of Trees through Image Segmentation Model*

You only look once (YOLO) [4] is a state-of-the-art, real-time object detection and segmentation system, based on convolutional neural network. We have taken Pytorch based YOLOv5 segmentation model[5], which has been widely used for segmentation purpose in digital imageries. The open source segmentation model is trained on 80 classes. Since Green Canopy Coverage calculation project is only focused on segmenting the green canopies, one needs to train the model with suitable dataset with canopy class. Here, we have taken a similar image dataset from Roboflow Universe, namely Canopy Segmentation Computer Vision Project [6] data. The dataset contains 388 annotated aerial imagery data, which have been distributed between train data (328 images) and test data (60 images).

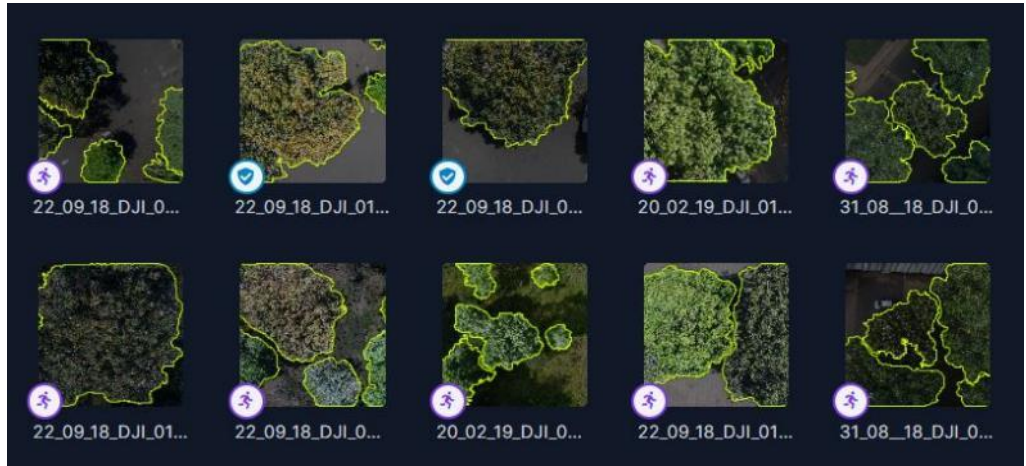

*Figure 7. Sample Green Canopy Cover Segmentation Dataset Images*

Thereafter, the YOLOv5 segmentation model has been trained with the stated dataset and onto shown desired accuracy on test data, we finalized the customized model, which can be treated as green canopy coverage segmentation model. The model typically takes 640 * 640 RGB images as input. While giving the image in different dimension, it can resize it in the stated dimension, but for achieving the best result, it is recommended to provide the input image in desired shape. Since the customized model has been trained with only one class, namely canopy, the model will be detecting the canopy class objects and segments them while execution. Finally, it yields a mask output, where the highlighted region is the canopy cover. If we overlay the mask onto the input image, we can visualize the segmented region as it is shown in the following image.

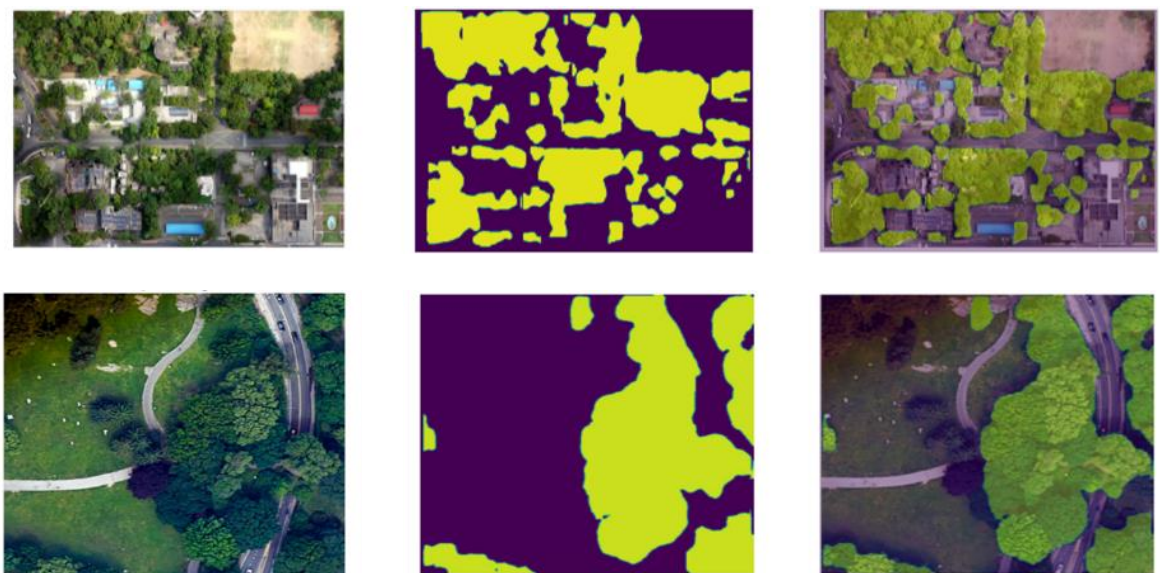

*Figure 8. [L] Original Input Image [M] Generated Segmentation mask [R] Overlayed Output Image*

## B. Object Detection Approach

Object Detection is a computer vision task in which the goal is to detect and locate objects of interest in an image or video. The task involves identifying the position and boundaries of objects in an image and classifying the objects into different categories. The model typically outputs bounding box coordinates, which are basically rectangular boxes to cover the detected objects in the image. Once the bounding boxes are generated, one can use the computer vision libraries like OpenCV-Python to create the bounding boxes over the input image and display them, varying in colors for different classes. The following image dictates how an object detection model can work on forest image and provide the canopies covered by boxes.

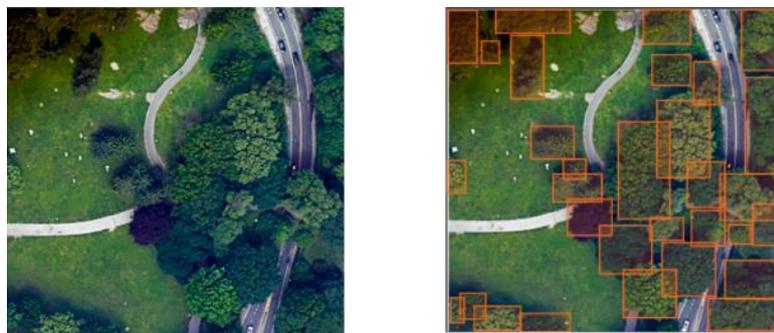

*Figure 9. Ideal Detection of Trees through Object Detection Model*

Like Segmentation approach, Pytorch based YOLOv5 object detection model[4] has been widely used for detection purpose in digital imageries. The open source object detection model is trained on 80 classes, same as segmentation model. Since Green Canopy Coverage calculation project is only focused on detecting the green canopies, one needs to train the model with suitable dataset with canopy class. Here, we have taken a similar image dataset from Roboflow Universe, namely Tree-Top-View Computer Vision Project[7] data. The dataset contains 580 annotated aerial imagery data, which have been distributed between train data (500 images) and test data (80 images).

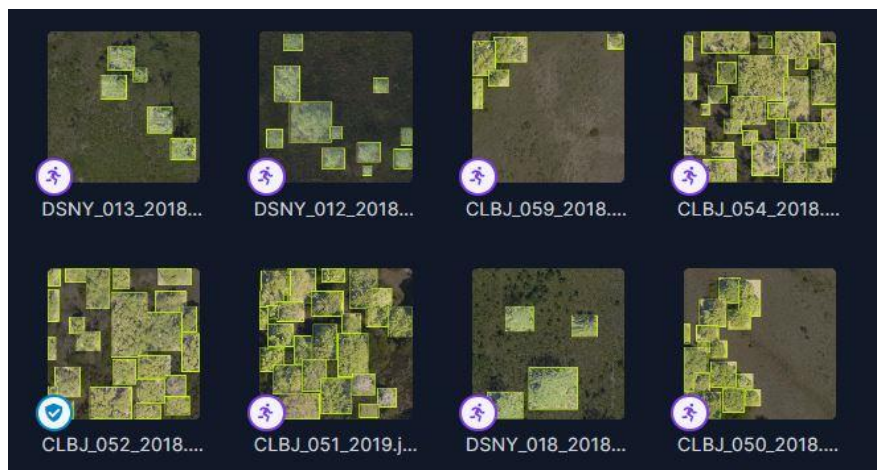

*Figure 10. Sample Green Canopy Cover Object Detection Dataset Images*

Thereafter, the YOLOv5 object detection model has been trained and generated the customized model, which can be treated as green canopy coverage detection model. Like the segmentation model, this model also typically takes 640 * 640 RGB images as input. Nonetheless, it is recommended to provide the input image in desired shape. Since the customized model has been trained with only canopy class, the model will be detecting the canopies and creates the boxes around them while execution. Finally, it yields a mask output, where the highlighted region is the canopy cover. If we overlay the mask onto the input image, we can visualize the detected region as it is shown in the following image.

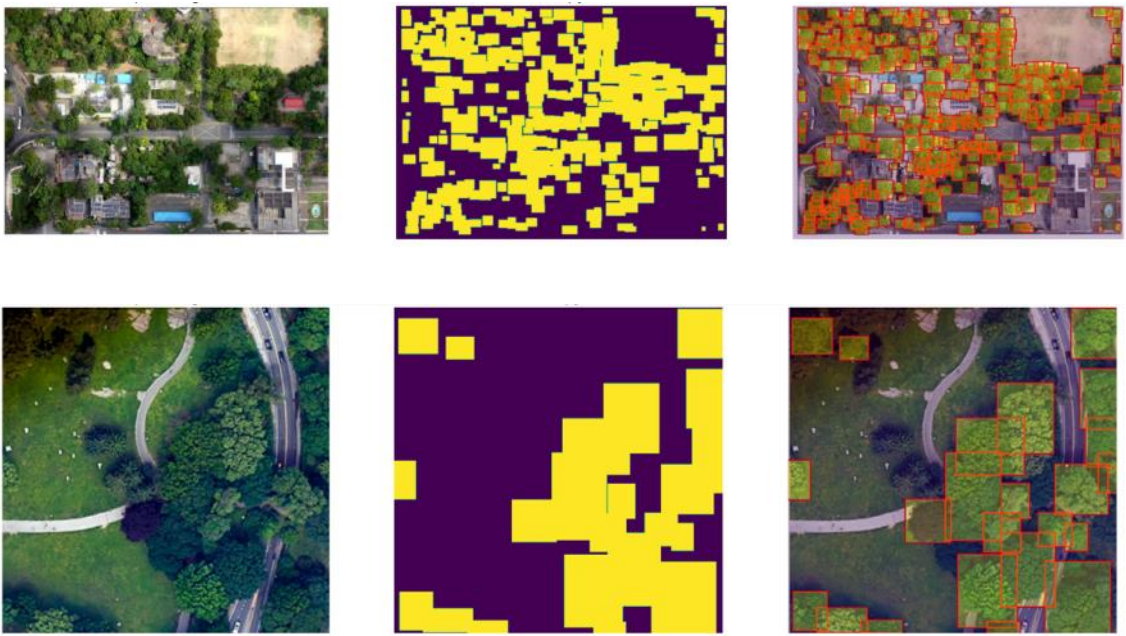

*Figure 11. [L] Original Input Image*
*[M] Generated Tree Canopy Detection Mask [R] Overlayed Output Image*

### C. Comparative Study

An aerial imagery of 23.46 acres of land along with the validated ground truth data were provided to us. The validation result (44.47%) aligns closely to the model outputs from object detection model (44.02%) and segmentation model (44.6%). The verification result necessitates a comparative study between the two models executed to select the optimal approach for estimating urban canopy coverage. While both models demonstrate promising result, a through comparison of their performance for different scenarios is crucial to determine which model excels in detecting and /or segmenting canopy areas.

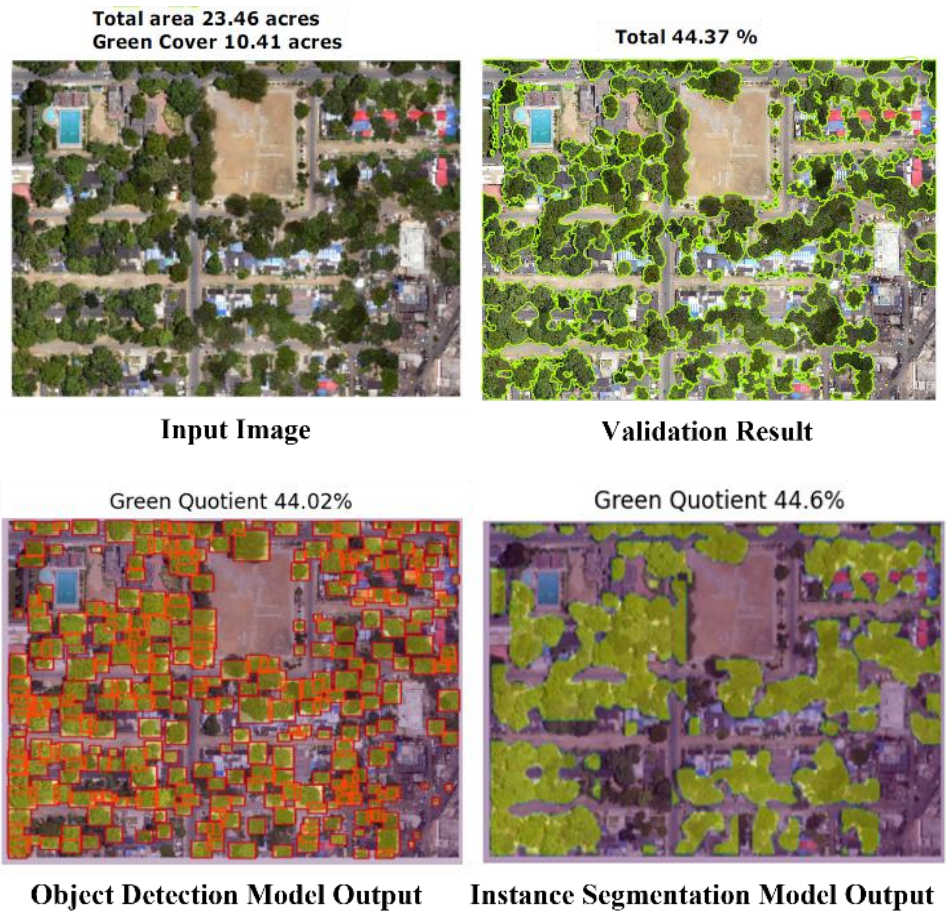

Total area 23.46 acres
Green Cover 10.41 acres

Total 44.37 %

**Input Image**

**Validation Result**

Green Quotient 44.02%

Green Quotient 44.6%

**Object Detection Model Output**

**Instance Segmentation Model Output**

*Figure 12. Comparison of generated output along with validation result*

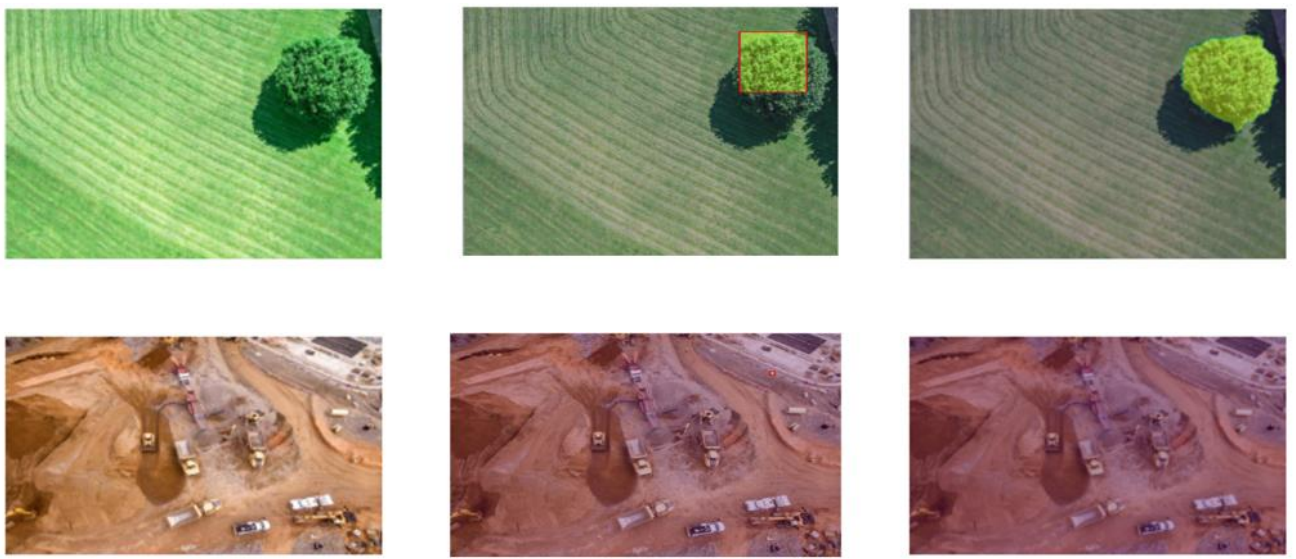

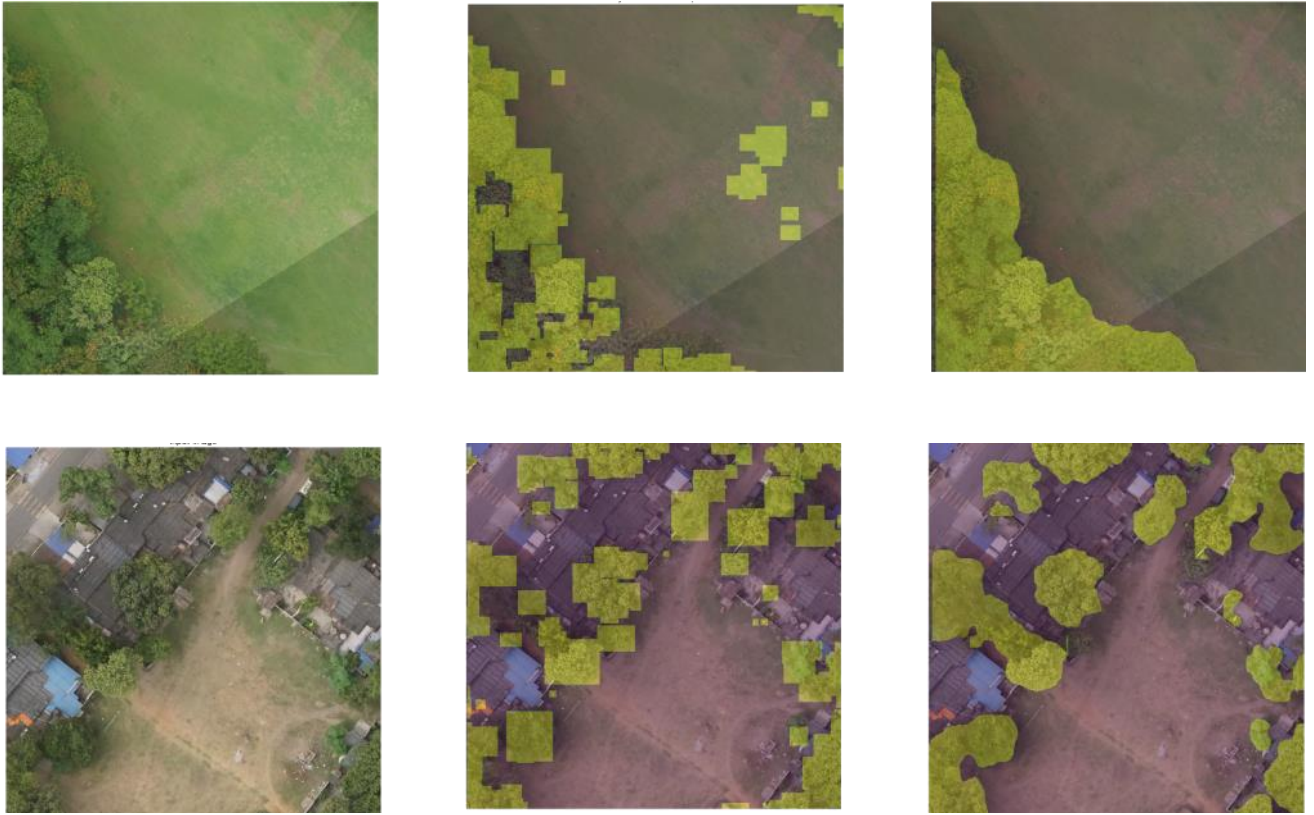

*Figure 13. Comparison of generated outputs by segmentation and detection models in different scenarios – single tree scenario, without canopy scenario, forest-grassland cooccurrence, typical urban image.*
*[L] Input Image [M] Object Detection Output [R] Segmentation Output*

Segmentation model is indeed providing better pixel level accuracy and better management of overlapping objects. Moreover, Segmentation doesn't rely on bounding boxes, which can miss smaller canopies and merged multiple objects, potentially underestimating canopy coverages. Canopy coverage is a spatially continuous phenomenon, and segmentation captures this more effectively. Segmentation provides a better precision of estimation for canopy coverages majorly because it classifies each pixel as canopy or non-canopy objects. Also, the rectangular shape of bounding boxes doesn't align with the mainly circular, continuous canopy top-view map. Therefore, for the final product, we proceed with customized Segmentation model for green canopy coverage calculation.

### D. Challenges and Mitigation Steps

- *Data Size and Format.*

    The gigantic size and intricate format of TIFF images pose substantial computational challenges in processing and analysis. This is further compounded by

the presence of multiple information layers within these files. Moreover, the to-go-library for accessing Tiff files, namely GDAL was unable to process the data. To overcome this hurdles, GeoTIFF library proves invaluable. Additionally, pre-processing techniques, specifically extracting pertinent information like the image channel, effectively minimize the data volume and processing time.

- *Data Transfer and Sharing.*

  Transferring large amounts of data between different cloud platforms like OneDrive and Google Cloud Storage (GCS) is indeed challenging due to factors such as limited data transfer rates, security concerns, and incompatibility issues. However, scheduling transfers during off-peak hours and continuous manual supervision helped overcoming these challenges and ensure efficient and secure data transfer.

- *Lack of Ground Truth data.*

  Without accurate ground truth data, it's challenging to assess the accuracy of canopy cover estimates obtained from deep learning methods performed over drone imagery. Limited ground truth data restricts the ability to train these algorithms effectively, potentially leading to inaccurate or biased predictions. Moreover, available relevant dataset over internet was significantly small in size. Augmentation techniques for images can help bypassing this challenge.

- *Computational Resource.*

  Analyzing canopy cover across extensive areas involves processing substantial datasets comprising numerous images. And for each chunk of the data, we are addressing CNN based methods. These algorithms often involve computationally intensive tasks like matrix operations, convolutions, and non-linear function approximations, due to which significant computational resources are required. We have used enterprise version of Google Colab along with GCS Bucket for storage purpose, which optimized our coding overheads. Although one must remember, while using cloud computing platforms can address computational limitations to some extent, it introduces considerations like network bandwidth, data transfer times, and potential cost implications.

- *Processing Time.*

  Processing large datasets often requires high computational power and specialized software, leading to increased resource allocation and potentially higher costs. Also, manual interventions in multiple steps works as a positive catalyst for lengthy processing times, which leads to delays in obtaining canopy cover data, delaying decision-making, and impacting project timelines.

## IV. Canopy Coverage Final Results

Area Covered out of Total Area **97.91%**
Green Canopy Cover Area **27.1%** according to Segmentation model

| | fileName | totalPixels | coveredPixels | segmentationPixels | coverPercentage | segmentationPercentage |
|---|---|---|---|---|---|---|
| 0 | tiles_A1 | 828993608 | 805825983 | 207039870 | 97.21 | 25.69 |
| 1 | tiles_A2 | 744243200 | 728376716 | 234227055 | 97.87 | 32.16 |
| 2 | tiles_A3 | 534528000 | 520829738 | 114770426 | 97.44 | 22.04 |
| 3 | tiles_A4 | 261734400 | 238869853 | 63516343 | 91.26 | 26.59 |
| 4 | tiles_B1 | 531251200 | 506459873 | 145942301 | 95.33 | 28.82 |
| 5 | tiles_B2 | 1296793600 | 1291256063 | 296591367 | 99.57 | 22.97 |
| 6 | tiles_B3 | 1329971200 | 1329971200 | 377455942 | 100.00 | 28.38 |
| 7 | tiles_B4 | 1172275200 | 1154281912 | 268610694 | 98.47 | 23.27 |
| 8 | tiles_C1 | 409600 | 0 | 0 | 0.00 | NaN |
| 9 | tiles_C2 | 221991680 | 212491798 | 55063430 | 95.72 | 25.91 |
| 10 | tiles_C3 | 793804800 | 776495613 | 204207191 | 97.82 | 26.30 |
| 11 | tiles_C4 | 735641600 | 710077553 | 274924444 | 96.52 | 38.72 |

*Figure 14. Tiff file wise Green Canopy Cover Calculation Result along with Percentage*

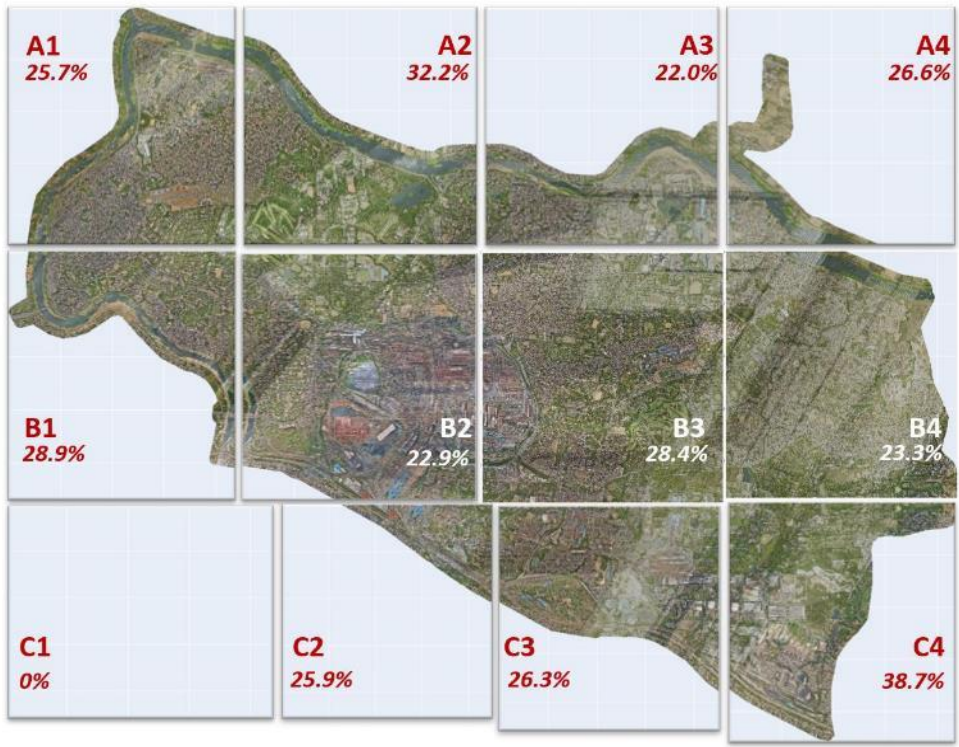

*Figure 15. Visualization of zone wise Green Canopy Cover Percentage of Jamshedpur*

## VI. Acknowledgements

Hereby extending sincere thanks to Krishna Prakash, Heat IT Transformation CS, for his exceptional coordination and management skills, which ensured the seamless execution of this project. We are also deeply indebted to Amar Pratap Singh, Assistant Manager Land Matters, for his diligent efforts in organizing and sharing the critical data required for our research. Consistent guidance and mentorship of Alok Kumar throughout the project were invaluable, providing insightful direction and support at every stage. We are also grateful to Krishna Ramachandran, Chief Analytics Officer, Tata Steel Ltd. and YN Rao and Moromee Das, Heads of Data & Analytics, Tata Steel Ltd., for their invaluable leadership and guidance. Their collective contributions were instrumental in achieving the project's goals.

**Appendix I. Green Canopy Coverage Gallery**

The resultant images are stored inside the Google Cloud Storage Bucket, inside the specified domain and the tabular data are stored in excel files in same location, inside folders corresponding to each tiff files. Demonstration of a couple of dozen sample customized green canopy coverage segmentation model generated results on 640 * 640 sized chunks of the data. Images are chosen randomly from all tiff data.

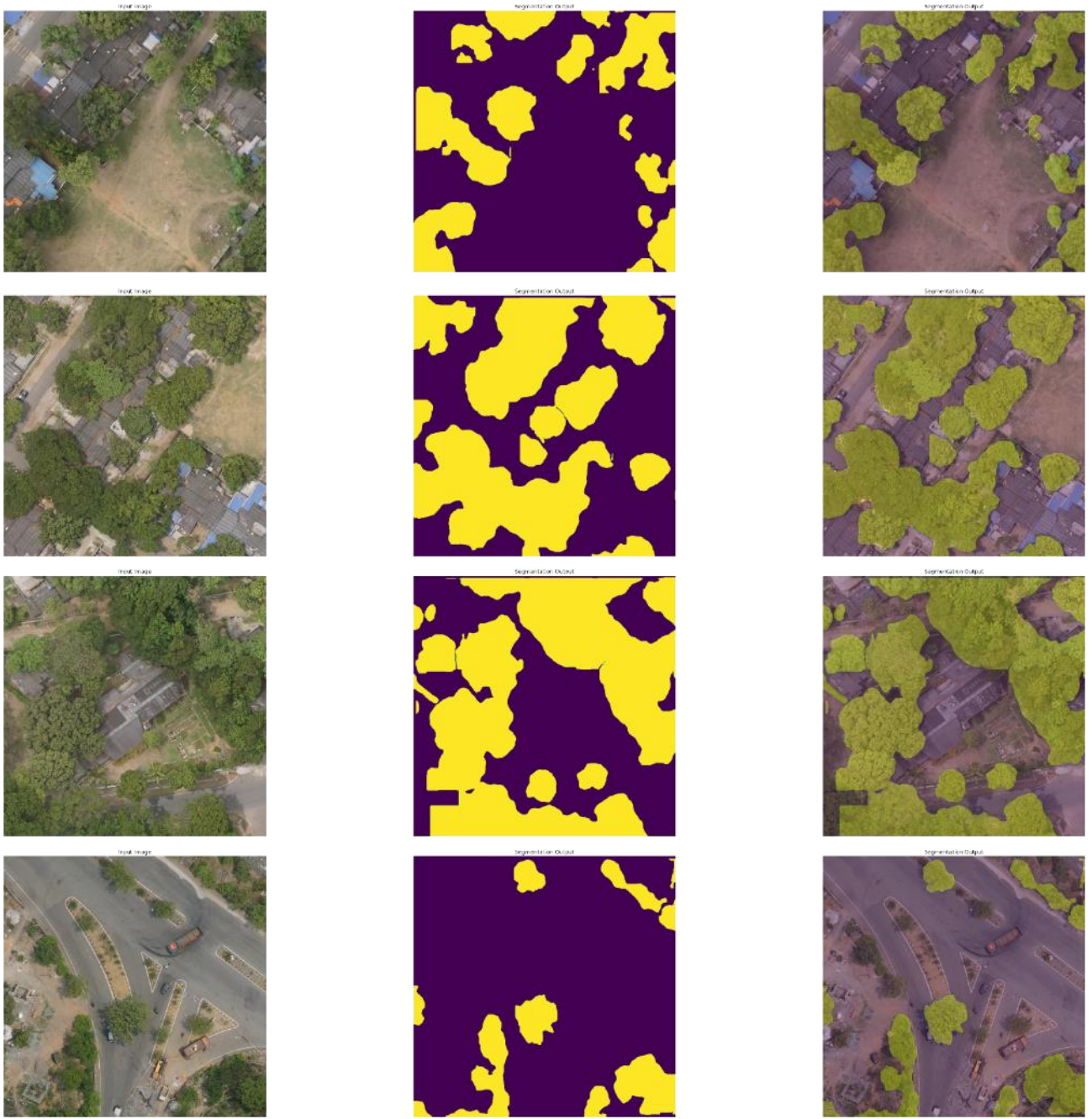

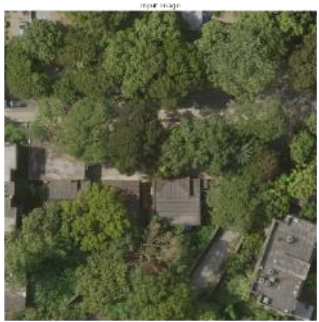 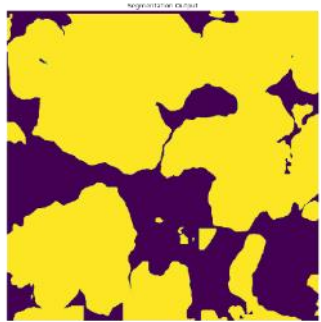 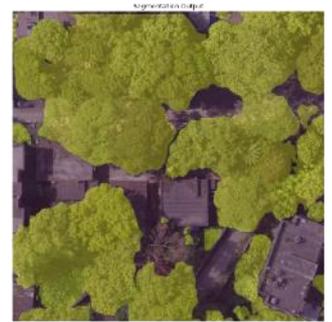

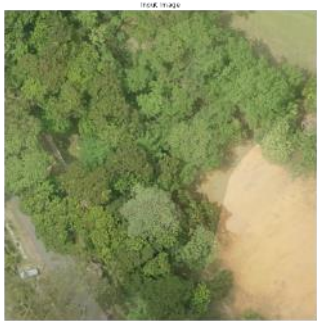 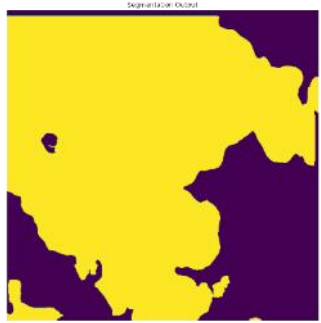 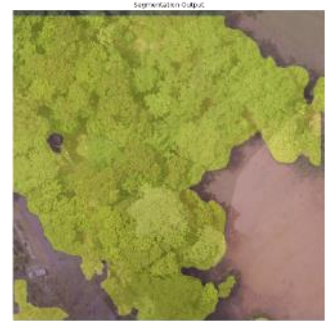

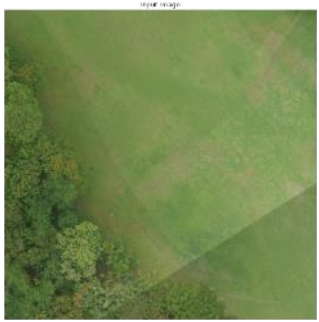 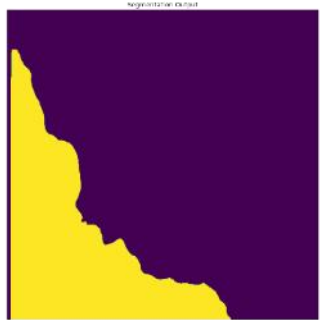 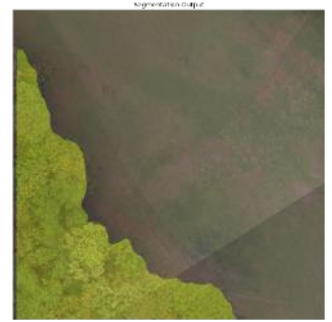

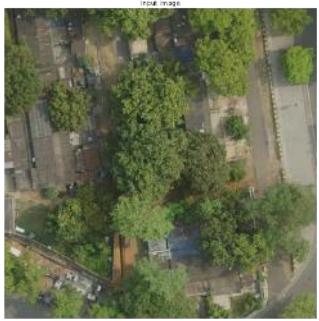 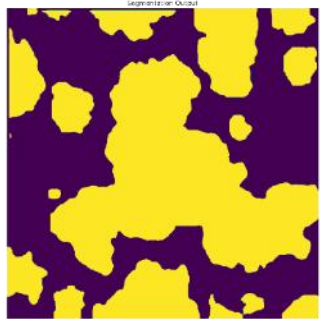 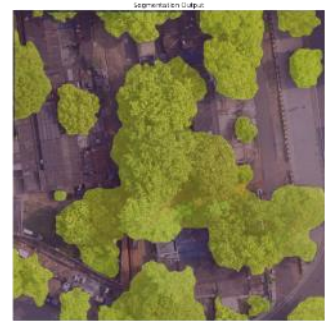

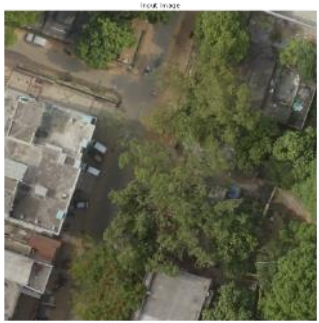 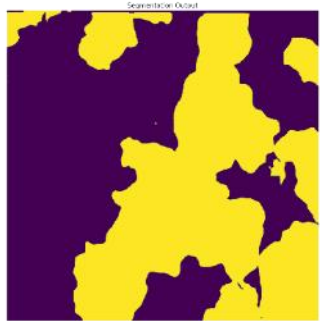 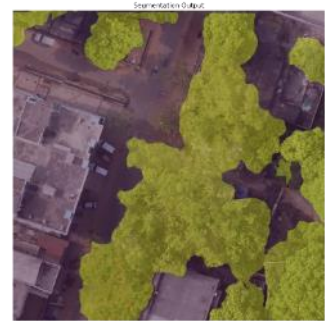

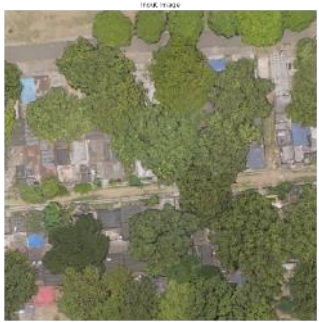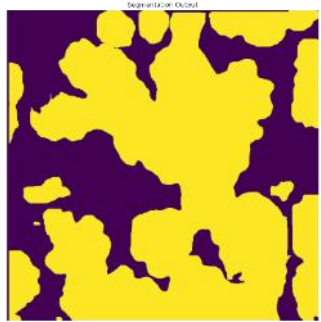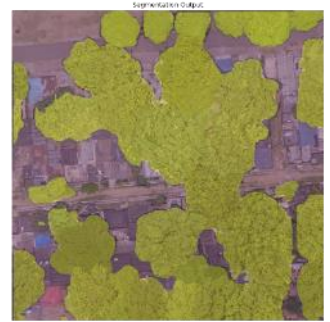

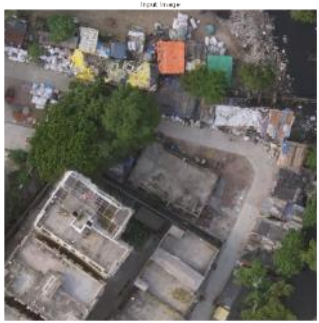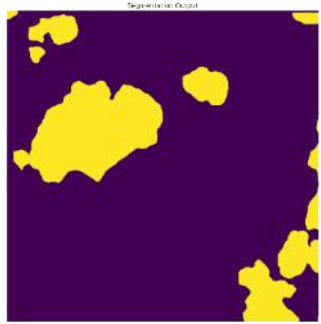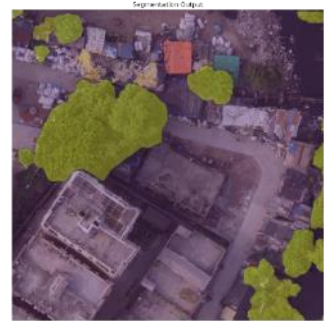

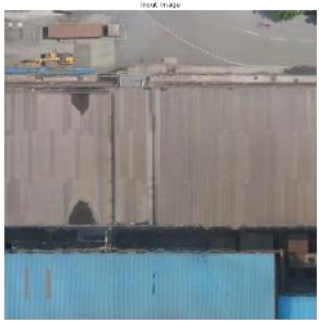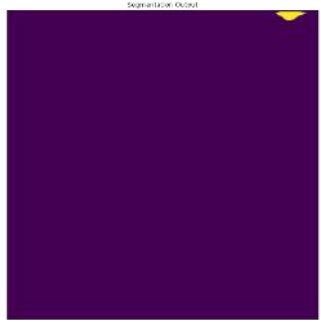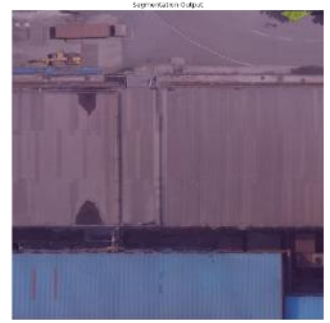

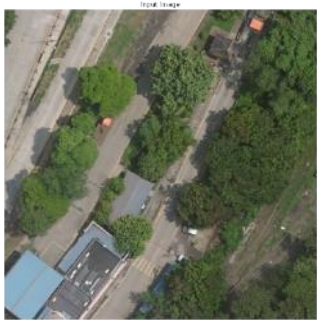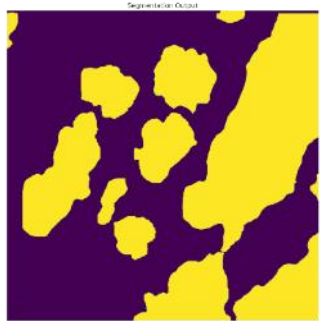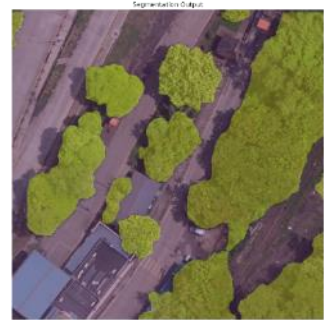

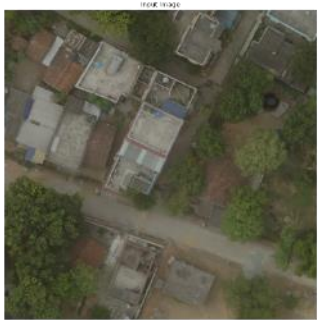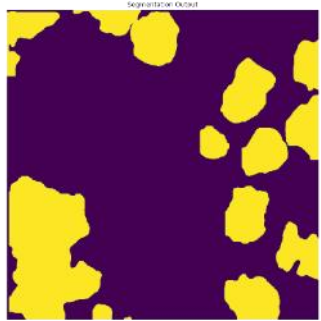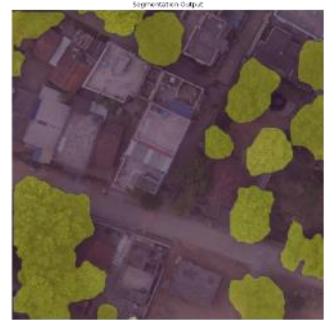

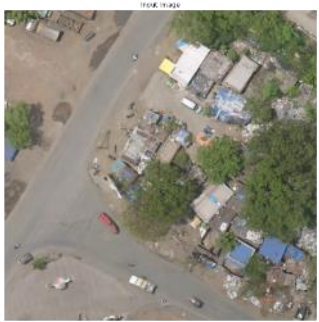
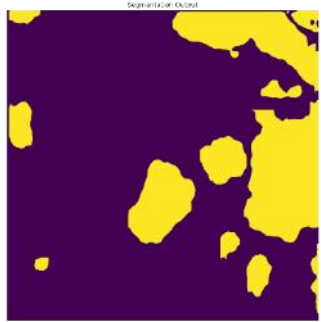
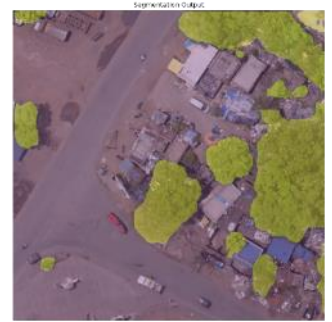

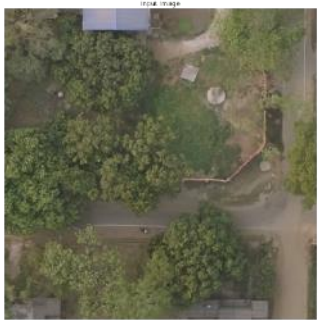
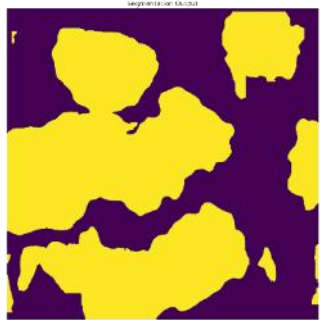
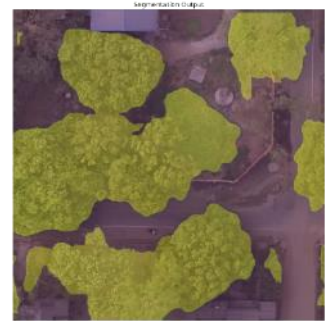

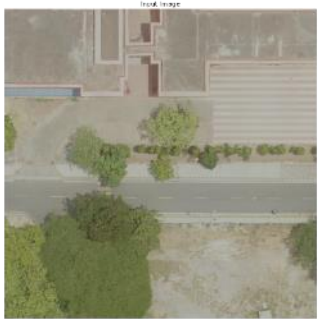
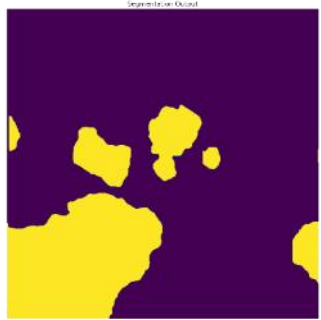
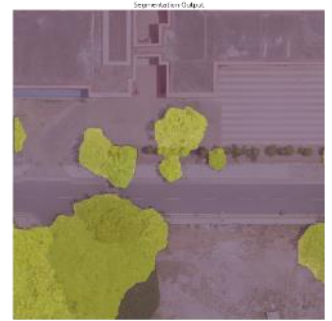

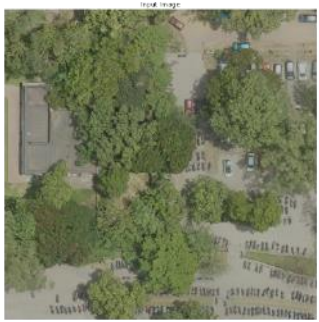
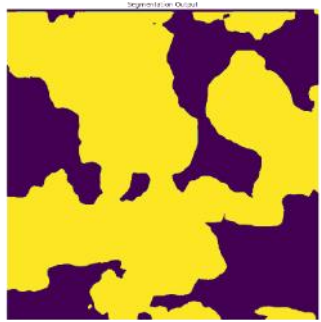
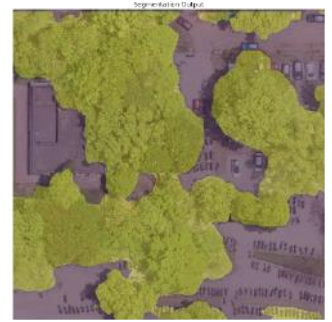

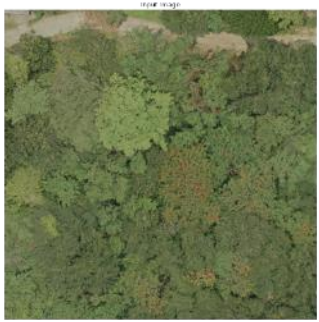
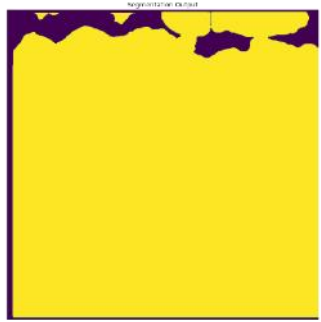
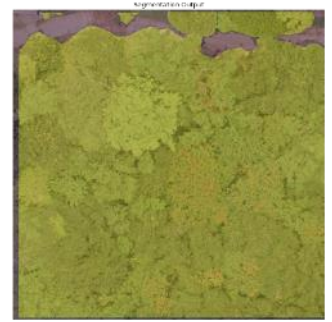

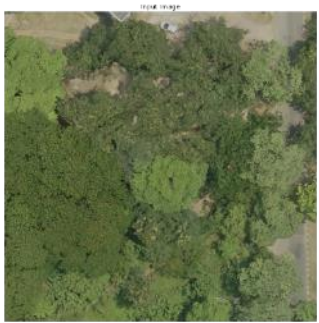 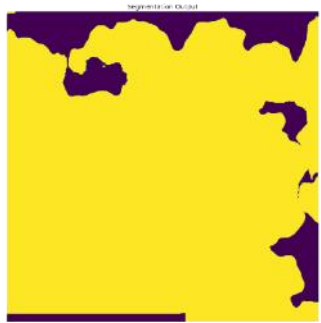 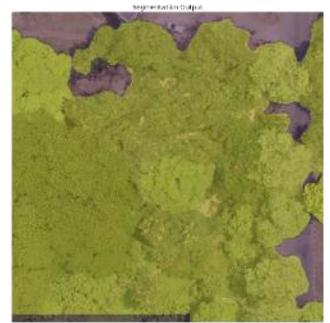

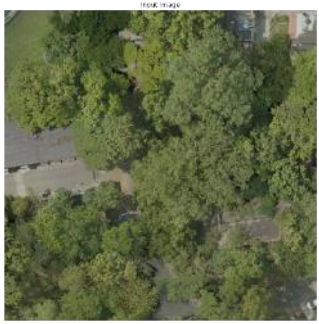 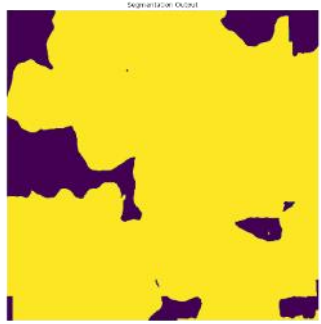 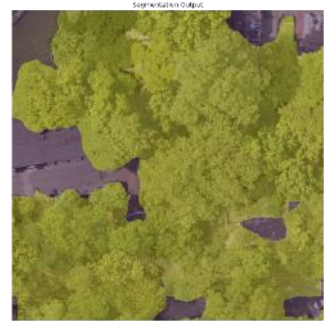

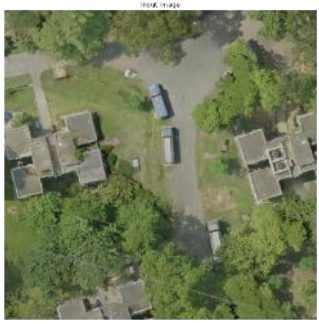 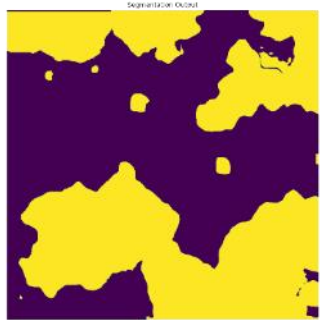 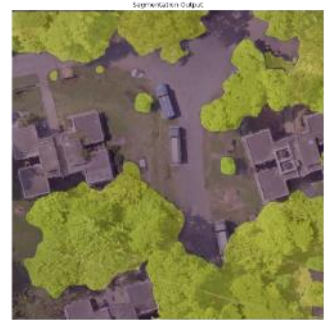

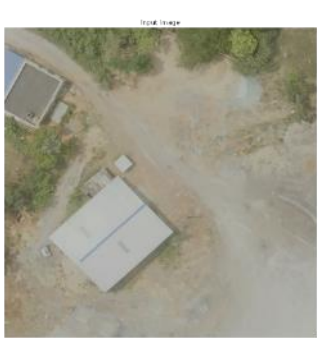 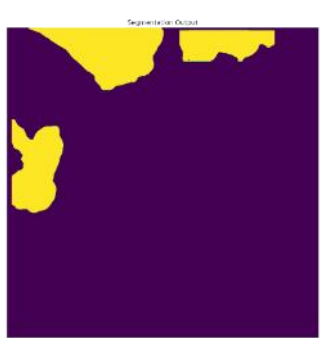 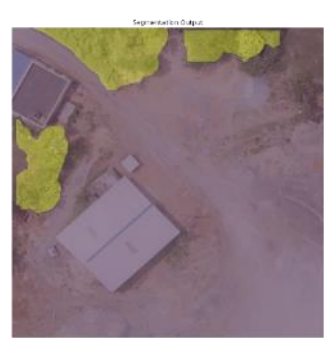

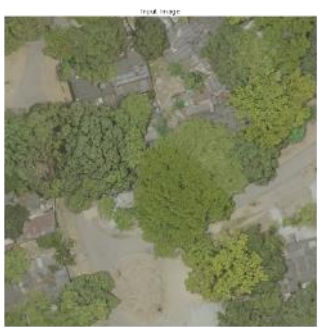 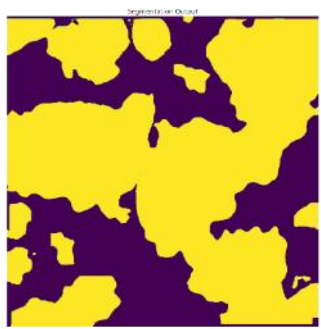 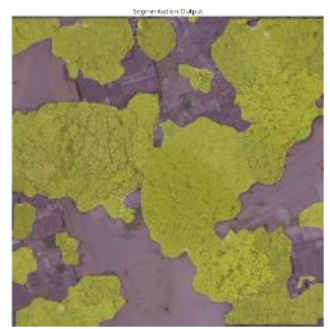